\pdfoutput=1

\documentclass[11pt]{article}

\usepackage[]{emnlp2021}

\usepackage{times}
\usepackage{amsmath}
\usepackage{latexsym}
\usepackage{multirow}
\usepackage{graphicx}
\usepackage{booktabs}
\usepackage{hyperref}
\usepackage{xcolor}
\usepackage{colortbl}
\usepackage{tabulary}
\usepackage{etoolbox}

\newcommand\base{\textsc{Base}}
\newcommand\ipa{\textsc{IPA}}
\newcommand\romani{\textsc{Romani}}
\newcommand\transl{\textsc{Inscrip}} 

\usepackage[utf8]{inputenc}

\usepackage{microtype}

\title{Alternative Input Signals Ease Transfer in Multilingual Machine Translation}

\author{Simeng Sun$^1$\hspace{3mm} Angela Fan$^2$\hspace{3mm} James Cross$^2$\\
{\bf Vishrav Chaudhary}$^2$\hspace{2mm}{\bf Chau Tran}$^2$\hspace{2mm} {\bf Philipp Koehn}$^2$\hspace{2mm} {\bf Francisco Guzmán}$^2$\\ 
University of Massachusetts Amherst$^1$ \hspace{1em} Facebook  AI$^2$\\
  \texttt{simengsun@umass.edu} \\
  \texttt{\{angelafan,jcross,vishrav,chau,pkoehn,fguzman\}@fb.com}
  
  }

\begin{document}
\maketitle

\begin{abstract}

Recent work in multilingual machine translation (MMT) has focused on the potential of positive transfer between languages, particularly cases where higher-resourced languages can benefit lower-resourced ones. 
While training an MMT model, the supervision signals learned from one language pair can be \emph{transferred} to the other via the tokens shared by multiple source languages. 
However, the transfer is inhibited when the token overlap among source languages is small, which manifests naturally when languages use different writing systems. 
In this paper, we tackle inhibited transfer by augmenting the training data with alternative signals that unify different writing systems, such as phonetic, romanized, and transliterated input. 
We test these signals on Indic and Turkic languages, two language families where the writing systems differ but languages still share common features. 
Our results indicate that a straightforward \emph{multi-source self-ensemble} -- training a model on a mixture of various signals and ensembling the outputs of the same model fed with different signals during inference, outperforms strong ensemble baselines by 1.3 BLEU points on both language families. 
Further, we find that incorporating alternative inputs via self-ensemble can be particularly effective when training set is small, leading to +5 BLEU when only $5\%$ of the total training data is accessible. 
Finally, our analysis demonstrates that including alternative signals yields more consistency and translates named entities more accurately, which is crucial for increased factuality of automated systems.

\end{abstract}

\section{Introduction}

Machine translation has seen great progress, with improvements in quality and successful commercial applications.
However, the majority of this improvement benefits languages with large quantities of high-quality training data (high-resource languages). 
Recently, researchers have focused on the development of multilingual translation models~\cite{aharoni-etal-2019-massively,fan2020englishcentric} capable of translating between many different language pairs rather than specialized models for each translation direction.
In particular, such multilingual models hold great promise for improving translation quality for \textit{low-resource} languages, as grouping languages together allows them to benefit from linguistic similarities as well as shared data between related languages. 
For example, training a translation system with combined Belarusian and Russian data would enable transfer learning between the two languages. 

We investigate how to enable multilingual translation models to optimally learn these similarities between languages and leverage this similarity to improve translation quality. 
The fundamental unit representing lingual similarity is the token --- languages that are similar often have similar words or phrases --- and during training, translation models can learn strong representations of tokens in low-resource languages if they are also present in high-resource languages.
However, a challenge arises when similar 
languages share only a small amount of tokens, 
which inhibits the transfer with limited and trivial cases of token sharing, e.g., punctuation marks and digits. 
This is particularly illustrated in cases where similar languages are written in different scripts, as the amount of shared tokens is small compared to languages using the same written script. 
An example would be Hindi and Gujarati, which have phonetic similarity but are written in their own native scripts.


\begin{figure*}
    \centering
    \includegraphics[width=0.8\textwidth]{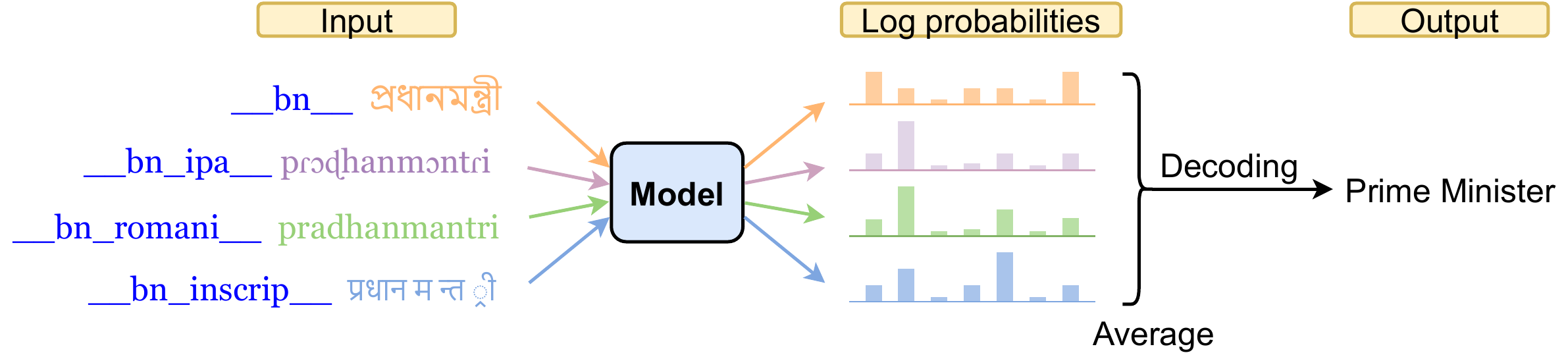}
    \caption{A generic illustration of self-ensemble for a multilingual translation system while translating Bengali to English. The input contains different signals, each preceded by a special language token (`\textcolor{blue}{\_\_bn\_\_}' indicates input in original Bengali script, `\textcolor{blue}{\_\_bn\_ipa\_\_}' the phonetic version of the same Bengli input, `\textcolor{blue}{\_\_bn\_romani\_\_}' the romanized version and `\textcolor{blue}{\_\_bn\_inscrip\_\_}' the same input but written in the script of Hindi, a language within the same language family). The log probabilities output by the model given each type of input are averaged for subsequent decoding process.}
    \label{fig:figure-1}
\end{figure*}

To tackle inhibited transfer due to distinct writing systems, we transform the original input via \emph{transliteration}, the process of converting text from one script to another, to get alternative signal from the original source sentences. 
Transliteration has been used in many real world cases, such as converting Cyrillic Serbian to Latin Serbian, as the language is commonly written with both scripts, or typing in romanized Hindi for convenience on a Latin-script keyboard.
To unify various writing scripts to increase token overlap, we experiment with three types of transliteration: \textbf{(1)} transliterate into phonemes expressed by international phonetic alphabet (\ipa), \textbf{(2)} transliterate into Latin script (\romani), and \textbf{(3)} transliterate into a script used by another language within the same language family (\transl). 
Beyond training on alternative inputs created through transliteration, we also systematically examine approaches to combining different signals.
Our experimental results on Indic and Turkic datasets demonstrate that \textbf{(i)} a \emph{self-ensemble} (Figure~\ref{fig:figure-1}) -- training a model on the mixture of different signals and using an ensemble of the \emph{same} model given different input signals during inference time, outperforms other methods such as multi-source ensemble and multi-encoder architecture, which require training multiple models or significant architectural changes. 
\textbf{(ii)} Further, without the need for additional bitext, a self-ensemble over the original and transliterated input consistently outperforms baselines, and is particularly effective when the training set is small (e.g. low-resource languages) with improvements of up to +5 BLEU. 
\textbf{(iii)} Finally, the improvements in BLEU originate from clear gain in the accuracy and consistency in the translation of named entities, which has strong implications for increased factuality of automated translation systems.


\section{Method}

Multilingual translation models enable languages to learn from each other, meaning low-resource languages can benefit from similarities to high-resource languages where data is plentiful.
However, surface-level differences between languages, such as writing system, can obscure semantic similarities. 
We describe an approach to transliterating input sentences to various alternative forms that maximize transfer learning between different languages, and various modeling approaches to incorporating such varied inputs.

\subsection{Alternative Inputs Bridge the Gap between Surface Form and Meaning}

While training a multilingual translation system, tokens shared by multiple source languages serve as anchors to transfer information obtained from learning one language pair to the other. 
For example, the translation of `terisini' in low-resourced Uzbek data can benefit from the word `derisinin' in relatively high-resourced Turkish data after tokenizing into sub-word units.
However, the transfer is hindered when the amount of shared tokens is small --- exacerbated by cases where the source and target languages are written in different scripts.
To alleviate the issue of various writing systems and encourage languages to transfer, we focus on alternative signals that unify the script of source languages and have larger token overlap. 
The core concept we explore is how to best leverage \textit{transliteration}, or the process of converting the text from one script to the other. 
We demonstrate that transliteration can be an effective data augmentation approach that improves the translation performance \emph{without the need of acquiring additional parallel data.} 
We explore three alternative inputs that allow models to share information more easily across languages with low token overlap but high semantic similarity. 
Figure~\ref{fig:example_signals} in Appendix~\ref{sec:appendix-signal-example} shows example alternative signals of the same Oriya sentence. 

\paragraph{Phonetic Input.} Related languages in the same language family usually sound similar, such as languages in the Romance language family and those in the Indo-Aryan language family. Although cognates can be captured to some degree for Romance languages on subword-level, it is difficult for the Indo-Aryan family as those languages use different writing systems. 
Therefore, to fully exploit shared information, we transform the original textual input (\base) into the phonetic space, where the basic units are phonemes expressed in international phonetic alphabet (\ipa). 
For example, \includegraphics[width=4em]{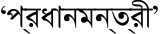} in Bengali looks like \includegraphics[width=6em]{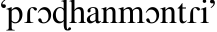} in IPA form.

\paragraph{Romanized Input.} Many languages use Latin alphabet (or Roman alphabet) in their default writing system, if not, they more or less have romanization of their default script in order to accommodate conventional keyboards, e.g., Chinese can be typed on U.S. keyboards through Pinyin, the romanization of Chinese. To utilize this existing form of alternative input, the romanized input is another signal we explore in this work. 
For example, \includegraphics[width=4em]{figures/bengali.pdf} looks like `pradhanmantri' in romanized form.

\paragraph{In-family Script Input.} 
The two previous alternative representations introduce tokens not present in the existing vocabulary, which increases the number of input and output representations the translation models must learn.
Further, phonetic input is artificial in the sense that it is not used by people to communicate to each other in written form --- and only used for pronunciation. 
Romanization naturally would introduce many additional tokens if the source language does not use Latin script. 
A third alternative that does not suffer these drawbacks is transliterate source language into the script of any of the other source languages in the multilingual translation model. To take advantage of language relatedness, we unify the source languages with the script used by a language within the same language family (\transl).
This method has the additional advantage of not needing to learn new subword tokenization models or replace the old vocabulary with a new one since all the inputs are expressed in one of the existing multilingual model's source language scripts.
For example, \includegraphics[width=4em]{figures/bengali.pdf} looks like \includegraphics[width=4em]{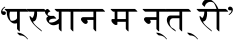}
when transliterated into Hindi script.

\paragraph{Advantages of Transliterated Inputs.} 

Various different input representations have been inserted into translation models, from parse trees~\cite{li-etal-2017-modeling,currey-heafield-2018-multi} to pretrained embeddings~\cite{artetxe2018unsupervised,conneau2018word}. 
Compared to these alternatives, transliteration has several clear advantages. 
Most importantly, transliteration is fast and accurate. 
Several existing alternatives often use other models to produce a different input, such as a parse tree, which cascades error from the first model into the translation model.
Comparatively, the alphabet alignment between various writing systems is quite well known, even for many low-resource languages, as alphabet is one of the foundational aspects of studying any new language. 
Similarly, phonetic pronunciation guides are often widely available.
These resources are also easily accessible programmatically, making them ideal for converting large quantities of supervised training data, for instance, the \texttt{espkea-ng} tool supports phonemization of more than 100 languages and accents. 
Beyond the ease of creating transliterations, we emphasize that this technique does not require any data annotation or collection of parallel data. 
Thus, it can be utilized in any existing translation system.

\subsection{Adding Transliterated Input Combinations to Translation Models} \label{sec:input-combination}

How can additional transliterated inputs be incorporated into modern machine translation architectures?
Since each alternative signal could capture a different view of the original input, in addition to training on each of the individual alternative signal alone, we investigate different approaches to combining them.

\paragraph{Straight Concatenation} The simplest combination strategy is to concatenate different input signals and separate them by a special token. For instance, to combine the original and phonetic input, we re-arrange the input to be of the format:  ``\textit{[original input]} \texttt{[SEP]} \textit{[phonetic input]}''.
During training, the decoder explicitly attends to tokens in both input signals. The advantage of this method is that no architectural change is required as all modification is operated on the input data. However, as the concatenated input becomes longer, this method requires more computation to train compared to the baseline model trained on the original input only.

\paragraph{Multi-Encoder Architectures} 

Prior works have found multi-encoder architecture to be effective for multi-source machine translation~\cite{nishimura-etal-2018-multi}. To cope with input from different sources, each encoder in the multi-encoder architecture deals with one type of input. To attend to multiple encoders on the decoder side, four cross-attention mechanisms can be adopted. We direct the readers to Appendix~\ref{sec:appendix-multi-enc} to get detailed description of these attention variations. Although prior works claim the efficacy of this approach, it is a complicated model choice requiring non-trivial architectural changes.

\paragraph{Multi-Source Ensemble} 
Ensembles are usually employed to boost the performance of a translation system. In a standard setup, each ensemble component is trained with identical configuration except for the random seed. 
We generalize this method to multi-source ensemble, i.e., individual ensemble components are trained on different transliterated inputs. During inference time, each component is fed with the type of transliteration it was trained on and produces the predicted log probabilities, which are averaged over all components for the subsequent decoding process. It is important for models trained on different source signals to have the same target vocabulary so that the average of log probabilities can happen. Unlike the previous two methods, this approach requires training multiple full models, thus requiring even more computation.

\paragraph{Multi-Source Self-Ensemble} 

Ensembling models that are trained on different input transliterations has the advantage that each individual model is maximally \textit{simple} --- only the input data for training changes. 
However, it comes with the downside that multiple different models need to be trained. 
This creates challenges particularly when models grow in size, as a new model would need to be created for each different transliterated input. 

Instead, we propose the \textit{Multi-Source Self-Ensemble}, which has all the advantages of traditional ensembling, but only requires \emph{one} model to be trained. 
Previous works in self-ensembles have focused on model robustness~\cite{liu2018towards}, which is distinct from varying input representations. 
Other work creates inputs in different languages~\cite{fan2020englishcentric}, but have to use a translation model to create those inputs first.

In our case, we train the model with different transliterated inputs mapping to the same translated target sentence. 
Concretely, the model is trained on the mixture of various input signals, each preceded by a special language token indicating which type of signal this input belongs to. 
At inference time, the alternative transliterated signals of the same test sentence are fed to the same model and the log probabilities produced by these separate passes are averaged as in multi-source ensemble.  
This approach is simple to implement as it requires no architectural change, meaning the transliterated inputs we propose can be added seamlessly to any existing translation library.
Unlike multi-source ensemble, only one model needs to be trained, stored and loaded for inference, greatly simplifying the ensembling process and increasing the scalability of our approach (particularly as translation models increase in size). To enforce fair comparison between multi-source self-ensemble and multi-source ensemble, we \emph{scale} the former so that it has the same number of parameters as that of all ensemble components of the latter. For the purpose of minimally impacting inference speed, the scaling is done only to the encoder embedding dimension so that the decoder remains the same.

\begin{table*}[]
\centering
\scalebox{0.95}{
\begin{tabular}{@{}lccl|llcc@{}}
\toprule
\multicolumn{3}{c}{$\sim$\textbf{93 M parameters}}  & & & \multicolumn{3}{c}{$\sim$\textbf{2$\times$93 M parameters}}  \\ \midrule
& \textbf{Indic} & \textbf{Turkic}  & & & & \textbf{Indic} & \textbf{Turkic}  \\ \midrule
\textbf{Single-input Original}  & \multicolumn{1}{l}{} & \multicolumn{1}{l}{} & & & \textbf{Standard Ensemble}  & \multicolumn{1}{l}{} & \multicolumn{1}{l}{} \\
\hspace{1.5em} \base  & 33.6  & 20.3  & & & \hspace{1.5em}\base+\base  & 34.5  & 21.1  \\ \midrule
\textbf{Single-input Alternative}  & \multicolumn{1}{l}{} & \multicolumn{1}{l}{} & & & \textbf{Multi-Source Ensemble} & \multicolumn{1}{l}{} & \multicolumn{1}{l}{} \\
\hspace{1.5em}\ipa & 32.7  & 17.9  & & & \hspace{1.5em}\base+\ipa  & 34.3  & 20.9  \\
\hspace{1.5em}\romani  & 32.5  & 20.7  & & & \hspace{1.5em}\base+\romani & 34.4  & 21.4  \\
\hspace{1.5em}\transl  & 33.4  & 20.5  & & & \hspace{1.5em}\base+\transl & 34.5  & 21.5  \\ \midrule
\textbf{Multi-Source Self-Ensemble} & \multicolumn{1}{l}{} & \multicolumn{1}{l}{} & & & \textbf{Multi-Source Self-Ensemble} & \multicolumn{1}{l}{} & \multicolumn{1}{l}{} \\
\hspace{1.5em}\base+\ipa  & 34.1  & 20.5  & & & \hspace{1.5em}\base+\ipa  & 35.7  & 21.9  \\
\hspace{1.5em}\base+\romani & 33.8  & 20.9  & & & \hspace{1.5em}\base+\romani & 35.7  & 22.2  \\
\hspace{1.5em}\base+\transl & \textbf{34.2}  & \textbf{21.3}  & & & \hspace{1.5em}\base+\transl & \textbf{35.8}  & \textbf{22.4}  \\ \bottomrule
\end{tabular}
}
\caption{BLEU scores on Indic test set and FloRes Turkic Devtest set.  }
\label{tab:main-res}
\end{table*}

\section{Experimental setup}

\paragraph{Dataset} We train our model on two language families: Indic and Turkic. The Indic dataset is from the WAT MultiIndic MT task\footnote{\url{https://lotus.kuee.kyoto-u.ac.jp/WAT/indic-multilingual/index.html}}, including 10 Indic languages and in total around 11 million Indic-English bi-texts. Six of the Indic languages are Indo-Aryan languages and the rest are Dravidian languages. All of these languages use a different writing system. The Turkic dataset is collected from the open parallel corpus~\cite{Tiedemann2012ParallelDT}\footnote{\url{https://opus.nlpl.eu/}}. For relatively high-resourced language Turkish, we randomly select 4 million subset from the CCAligned~\cite{elkishky_ccaligned_2020} corpus. Within this dataset, two languages use Cyrillic alphabet (Kazakh and Kyrgyz) and the rest use Latin alphabet. Detailed dataset statistics are displayed in Table~\ref{tab:data_stats} in Appendix~\ref{sec:appendix-experiments}.

\paragraph{Single-input model} To test the effectiveness of each input signal, we train models on each \emph{single} type of input: 
original input (\base), phonetic input (\ipa), romanized input (\romani) or input all expressed in the script of a language within the same language family (\transl).
On the Indic dataset, for the \transl\ signal, all Indo-Aryan languages are transliterated into Hindi script, and all Dravidian languages into Tamil script. On the Turkic dataset, all languages in Latin script are transliterated into Cyrillic script.

\paragraph{Multi-Source Ensemble} 
A baseline for ensembling models trained on different signals is the standard ensemble (\base+\base) where two \base\ models are ensembled, each trained with a different random seed. Although there are multiple combinations of input signals, we only discuss the cases where \base\ is combined with one of the alternative inputs (\base+\{\ipa,\romani,\transl\}), since in our preliminary experiments, we found dropping the \base\ model leads to significantly degraded performance.

\paragraph{Multi-Source Self-Ensemble} Similar to above, we train a single model on the mixture of original input and one of \{\ipa,\romani,\transl\} input for multi-source self-ensemble.
To enforce fair comparisons with the ensembled models, which have more parameters in total, we train two sizes of the self-ensemble (SE) model, one having the same size of a single baseline model, the other scaled to have twice the number of parameters of a single \base\ model. 

\paragraph{Data Preprocessing} We use \texttt{espeak-ng}\footnote{\url{https://github.com/espeak-ng/espeak-ng}} to convert the original input to phonetic input. For Indic languages, we use \texttt{indic-trans}\footnote{\url{https://github.com/libindic/indic-trans}}~\cite{Bhat:2014:ISS:2824864.2824872} to obtain the romanized as well as the in-family transliterated input. On the Turkic dataset, we manually align the Cyrillic and Latin alphabet and substitute the letter(s) in one script with the corresponding one in another.\footnote{The substitution process starts from the letter in the target script that corresponds to the most number of letters in the source script.} 
The Indic languages are tokenized with \texttt{indic\_nlp\_library} and the rest are tokenized with \texttt{mosesdecoder}\footnote{\url{https://github.com/moses-smt/mosesdecoder}}. We use \texttt{sentencepiece}\footnote{\url{https://github.com/google/sentencepiece}} to get 32K BPE~\cite{sennrich-etal-2016-neural} subword vocabularies.

\paragraph{Training \& Evaluation} We train many-to-En language directions during training (10 and 5 directions for Indic and Turkic dataset respectively). The architecture is a standard 6-layer encoder 6-layer decoder Transformer model, with 512 embedding dimension and 2048 hidden dimension in the default setting. For the scaled self-ensemble model, we increase the encoder hidden dimension such that the number of parameters in this model approximately matches that of $n$ baseline models ($n=2$ for results in Table~\ref{tab:main-res}). We use $4000$ warmup steps and learning rate $0.0003$. 
Both the dropout and attention dropout rate are set to $0.2$. Label smoothing is set to 0.1. Data from different language pairs are sampled with 1.5 temperature sampling. We train all models for 18 epochs and 40 epochs for Indic and Turkic dataset respectively and evaluate the best checkpoint selected by dev loss. We use \texttt{spBLEU}\footnote{\url{https://github.com/facebookresearch/flores\#spm-bleu}}~\cite{flores_1,flores_2} to compute the BLEU scores.

\section{Results} \label{sec:results}

In this section, we compare the performance of our proposed multi-source self-ensemble model to various alternative ways of input combinations on two low-resource language families: Indic and Turkic languages. Furthermore, we show multi-source self-ensemble learns faster and generates more consistent and accurate translations. 

\subsection{Performance of Multi-Source Self-Ensemble}

Our method is based on the hypothesis that incorporating alternative inputs increases the token overlap of source languages, which benefits the transfer during training. To verify this, we compute average sentence-level uni-gram overlap of all source language pairs (Table~\ref{tab:token_overlap}) and find that alternative signals do have higher token overlap compared to the original input. For instance, the \ipa\ signal, having similar average sentence length as \base\ , has much higher token overlap (0.15 vs. 0.03).

\begin{table}[]
    \centering
    \scalebox{0.88}{
    \begin{tabular}{l|cccc}
    \toprule
         & \base & \ipa & \romani & \transl \\\midrule
Uni-gram & 0.03 & 0.15 & 0.13 & 0.16\\
Sent. len & 34.7 & 39.3 & 25.9 & 51.3\\
    \bottomrule
    \end{tabular}
    }
    \caption{Uni-gram token overlap and sentence length of various types of input on MultiIndic dev set.}
    \label{tab:token_overlap}
\end{table}

Do increased token overlaps result in better translation performance? We train models on each of the alternative inputs alone and report the results in the left column of Table~\ref{tab:main-res}. 
We find that using only one alternative input in the source has either worse or similar performance as the original baseline, indicating higher token overlap among source languages does not guarantee better BLEU scores. The degraded performance is likely due to unfavorable interference introduced by shared tokens in the alternative signals. The interference may take the form of information loss\footnote{For example, the punctuation marks are lost during phonemization process.} or increased ambiguity\footnote{For instance, multiple words may have the same pronunciation and thus have the same input in \ipa\ form, which makes the learning harder.}, which reinforces the importance of combining alternative inputs with the original input.

Due to undesired interference exhibited in the alternative input spaces, 
we therefore adopt the input combination using our proposed \textit{Multi-Source Self-Ensemble} to combine the original input and alternative signals. Results in left lower part of Table~\ref{tab:main-res} demonstrate improvements over the single-input baseline. Our best performing alternative input configuration improves +1.0 BLEU on Turkic languages and +0.6 BLEU on Indic languages for 93M parameter models. 

In production, model ensembles are often employed to achieve the best possible performance. We also provide results against these strong ensemble baselines and observe +1.3 BLEU improvements on both Indic and Turkic languages. Note that, to enforce a fair comparison, we compare a scaled version of the multi-source self-ensemble model which has the same number of parameters as multiple ensemble baseline components.


\begin{table}[!t]
    \centering
   \scalebox{0.9}{
    \begin{tabular}{@{}lc@{}}
\toprule
\textbf{Configuration }                           & \textbf{BLEU} \\ \midrule
\textbf{Single-input Baseline}                   &      \\
\hspace{3em}\base                                     & 33.6 \\ \midrule
\textbf{Straight Concatenation}                  &      \\
\hspace{3em}\base+\texttt{<SEP>}+\ipa    & 33.7 \\
\hspace{3em}\base+\texttt{<SEP>}+\romani & 33.7 \\
\hspace{3em}\base+\texttt{<SEP>}+\transl & 33.6 \\ \midrule
\textbf{Multi-Encoder Architectures }         &      \\
\hspace{1em} \textbf{Bi-Encoder} & \\
\hspace{3em}\base+\base                                & 34.2 \\
\hspace{3em}\base+\ipa                                 & 33.9 \\
\hspace{3em}\base+\romani                              & 33.9 \\
\hspace{3em}\base+\transl                              & 34.0 \\
\hspace{1em} \textbf{Quad-Encoder} & \\
\hspace{3em}\base+\base+\base+\base                      & 34.3 \\
\hspace{3em}\base+\ipa+\romani+\transl                   & 34.1 \\\midrule
\hspace{1em} \textbf{Multi-source Self-ensemble} & \\
\hspace{3em}\base+\transl &  34.2 \\
\bottomrule
\end{tabular}
   }
    \caption{Indic test set BLEU of models trained on straight concatenation of input as well as multi-encoder architectures. Training on the concatenated input does not impact the BLEU much. Multi-encoder architectures, although having a lot more number of parameters, for instance, quad-encoder, achieve similar performance of a much smaller multi-source self-ensemble.  }
    \label{tab:not-working-methods}
\end{table}

\begin{figure}
    \centering
    \includegraphics[width=0.49\textwidth]{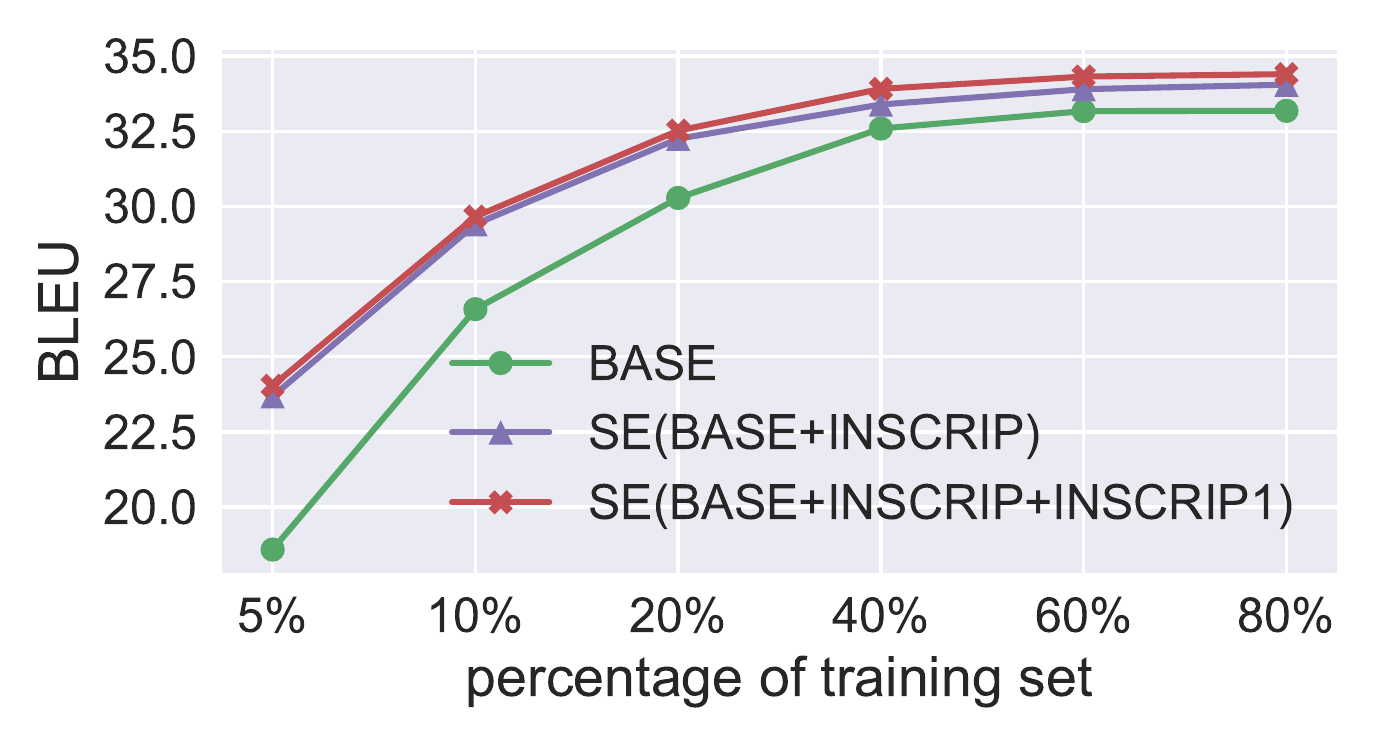}
    \caption{Learning curve of the baseline model \base\ and the same-sized self-ensemble model trained on the original input as well as transliterated input. \transl\ denotes the transliteration where the target script for Indo-Aryan and Dravidian languages are Hindi and Tamil respectively. The target scripts of \transl1\ are Oriya and Kannada respectively.
    }
    \label{fig:learning-curve}
\end{figure}

\subsection{Advantages of Multi-Source Self-Ensemble}

\paragraph{Architectural Simplicity.}

As introduced in \S\ref{sec:input-combination}, there are various ways to incorporate multiple inputs, such as concatenation to form a longer input or using multiple encoders networks.
In Table~\ref{tab:not-working-methods}, we show that using multiple encoders has no improvements over the comparable baseline with raw text input, and straight concatenation only brings marginal gains (+0.1 BLEU).
Further, our simple but effective Multi-Source Self-Ensemble technique reaches the same performance as that of a much larger quad-encoder model, which requires non-trivial architectural changes and takes more compute to train.
Thus, our technique is suitable to be used out of the box in any seq-to-seq library. 

\paragraph{Faster Learning in Low-Resource Settings.}

To understand how self-ensemble performs with different amounts of data, we plot the learning curve of both the baseline and the self-ensemble model on $5\%$\footnote{When the training set is very small ($5\%$ and $10\%$), we train for 60 epochs and select the model by dev loss.} to $80\%$ of the total Indic training set.\footnote{The transliterated input are those of the same subset of training data, thus no sentences having new semantic meaning are added in the multi-source self-ensemble setup.}
As shown in Figure~\ref{fig:learning-curve}, the self-ensemble model outperforms the baseline model by a large margin when the amount of training data is small (\textbf{+5} \textbf{BLEU} when only $5\%$ of the total set is used for training). 
This is the scenario for most low-resource languages, as the gap gradually closes when more data is available. 
Overall, the multi-source self-ensemble model is consistently better than the baseline model irrespective of training data scale. This suggests that transliteration can be a cheap and effective data augmentation approach when used in conjunction with multi-source self-ensemble.

\begin{table}[]
    \centering
    \small 
\scalebox{0.97}{
\begin{tabular}{lccl}
\toprule
  & \textbf{C-BLEU}  & \textbf{NE-F1} & \\ \midrule
\textbf{Single-input Baseline} &  & \\
\hspace{1.5em}\base  & 34.7 & 55.9 \\ \midrule
\textbf{Single-input Alternative Input} & \multicolumn{1}{l}{} & \multicolumn{1}{l}{} \\
\hspace{1.5em}\ipa  & 33.8 & 54.7 \\
\hspace{1.5em}\romani  & 33.0 & 54.5 \\
\hspace{1.5em}\transl  & 35.3 & 55.4 \\ \midrule
\textbf{Multi-Source Self-Ensemble} & \multicolumn{1}{l}{} & \multicolumn{1}{l}{} \\
\hspace{1.5em}\base+\ipa  & \textbf{36.2} & 56.1 \\
\hspace{1.5em}\base+\romani  & 35.5 & 56.3 \\
\hspace{1.5em}\base+\transl  & \textbf{36.2} & \textbf{56.4} \\ \bottomrule
\end{tabular}
}
    \caption{The consistency BLEU (\textbf{C-BLEU}) and exact named entity match F1 (\textbf{NE-F1}) of MultiIndic test set. Higher \textbf{C-BLEU} scores imply more consistent output in many-to-En setting. Higher \textbf{NE-F1} scores indicate better translation of named entities.  
    }
    \label{tab:analysis-res}
\end{table}

\paragraph{Improved Output Consistency.} \label{sec:consist-output}

We conduct a deeper analysis to understand the performance improvement of Multi-Source Self-Ensembles beyond BLEU scores alone.
We find that our proposed technique generates much more consistent output, which could be a benefit of alternative signals transferring information more easily amongst source languages. 
We propose consistency BLEU (\textbf{C-BLEU}) to quantify the consistency of multi-way evaluation output of a many-to-En translation model. 
We treat the output of $L_1$-En direction as reference and output of all other $L_i$-En directions as hypothesis. 
We compute this for all $N$ source languages in the dataset, accounting for total $N(N-1)$ C-BLEU scores, then take the average of all(Table~\ref{tab:analysis-res}). 
While training on \ipa\ or \romani\ alone does not outperform the baseline in terms of C-BLEU, model trained on \transl\ input improves the score by +1.3. Self-ensemble over \base\ and \ipa\ increases the C-BLEU to \textbf{36.2} (and from \textbf{36.3} to \textbf{38.1} with scaled model), indicating the alternative signals are best trained together with the original input. 

\begin{figure*}
    \centering
    \includegraphics[width=0.9\textwidth]{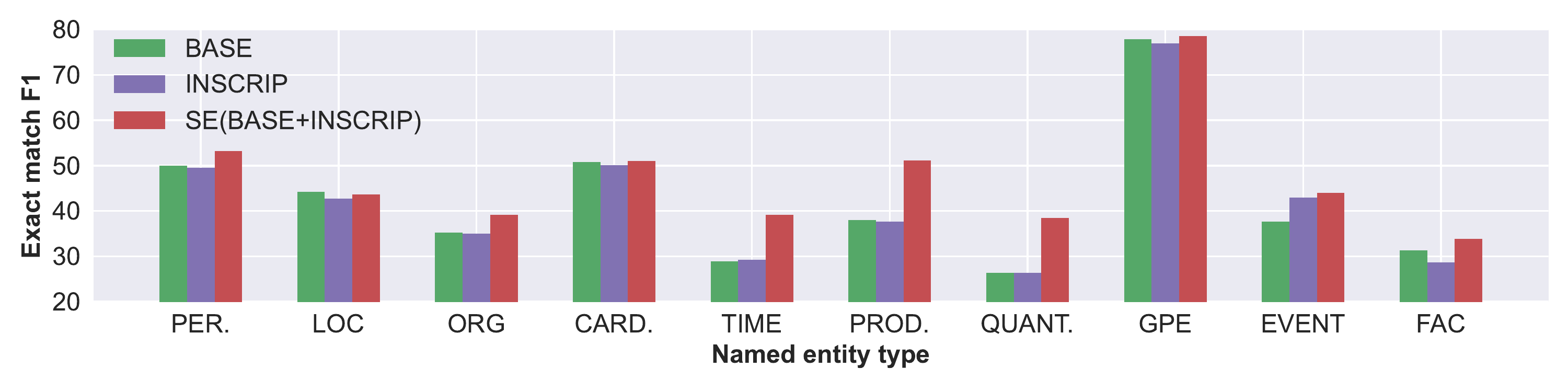}
    \caption{The exact named entity match F1 score of \base\, \transl\ and same-sized self-ensemble model trained on the previous two inputs (SE(\base+\transl)). Although the results of the self-ensemble model only slightly outperforms the baseline (55.9 vs. 56.4), the gains are more obvious when breaking the results by entity type. }
    \label{fig:named-entity} 
\end{figure*}

\paragraph{Improved Named Entity Accuracy.}

The previous analysis implies the self-ensemble model outputs more consistent translation, yet this does not mean the consistent translations are accurate. In this section, we conduct an analysis targeted at named entities. We use spaCy~\cite{spacy} NER tagger to extract all named entities, and then compute the exact match of the extracted entities. 
According to the results in Table~\ref{tab:analysis-res}, self-ensemble introduces small gains (+0.5) in terms of named entity F1 (NE-F1), whereas the scaled self-ensemble boosts NE-F1 score by \textbf{+1.1}. 
Although the improvement is small in aggregate, we find significant improvement when breaking down by entity type. 
As shown in Figure~\ref{fig:named-entity}, the multi-source self-ensemble model (without scaling) outperforms the baseline model on certain entity types, e.g., person, organization, time and event by a large margin.

\section{Related work}

\subsection{Alternative Input for Multilingual MT}

Our work can be viewed as multilingual MT~\cite{firat-etal-2016-zero} combined with multi-source MT~\cite{zoph-knight-2016-multi}, where the sources are not other languages but rather alternative transliterated signals. The transliterated input has been explored in the past for translation system. \citet{nakov-ng-2009-improved} use transliteration as a preprocessing step for their phrase-based SMT model to tackle systematic spelling variation.  Both ~\citet{chakravarthi_et_al:OASIcs:2019:10370} and ~\citet{koneru-etal-2021-unsupervised} convert  Dravidian languages to Latin script and train multilingual models with \emph{both source and target} in Latin script; the latter identify code-switching to be a challenge during back-transliteration. Besides converting to Latin script, ~\citet{dabre-etal-2018-nicts} use another common script, Devanagari, for Indic languages.
In addition to the natural written scripts, previous works also explored artificial script, such as IPA. ~\citet{liu-etal-2019-robust} incorporate phonetic representations, specifically for Chinese Pinyin, to cope with homophone noise. Unlike our work, ~\citet{chakravarthi_et_al:OASIcs:2019:10370} adopt transliteration to IPA for both the source and target. 
Apart from transliterated input, other potential alternative signals we did not fully explored include orthographic syllable units~\citep{kunchukuttan-bhattacharyya-2016-orthographic, kunchukuttan2020utilizing}, morpheme-based units~\citep{ataman2017linguistically,dhar-etal-2020-linguistically}, and character~\cite{lee-etal-2017-fully} or byte~\cite{wang2019neural} level input in addition to the subword-level units~\cite{sennrich-etal-2016-neural}.

\subsection{Input signal combination}

Multi-encoder architecture is the most common way to combine input from different sources. While previous works mainly use additional encoders to encode syntactic information~\citep{li-etal-2017-modeling,currey-heafield-2018-multi} or input in another language~\cite{nishimura-etal-2018-multi}, we feed in each encoder with different signals of the same sentence. Prior works also investigated approaches to combining input at different granularity~\cite{ling2015characterbased,chen-etal-2018-combining,casas-etal-2020-combining}. ~\citet{wang2018multilingual} combine the decoupled lexical and semantic representations through an attention mechanism. Another common method of utilizing additional input signal is multi-task learning, force the model to output extra labels~\cite{luong2016multitask,gronroos-etal-2017-extending}. Apart from combining the sources during training, inference-time ensemble~\cite{garmash-monz-2016-ensemble} is often adopted by recent submissions to shared MT tasks~\cite{ng2019facebook,tran2021facebook}. The ensemble components are usually separate systems trained with different random initialization or language pairs. ~\citet{fan2020englishcentric} ensemble the same model by feeding in source sentences in different languages. The self-ensemble approach was also found to make networks more robust after adding random noises~\cite{liu2018towards}. Prior work also uses the term "self-ensemble" to refer to an ensemble of models using weights from different time steps during training~\cite{Xu2020ImprovingBF}.

\section{Conclusion}

To overcome the low token-overlap issue exhibited in multilingual MT systems due to distinct writing system, we examined three alternative signals (phonetic, romanized and in-family transliterated input) and investigated four approaches (input concatenation, multi-encoder, multi-source ensemble, self-ensemble) to combining them with the original input. Our results show that training a single model with a mixture of diverse signals and performing self-ensemble during inference time can improve BLEU by 1.3 points on Indic and Turkic dataset. The improvements can reach +5 BLEU when training data size is small. Further, we show this approach generate more accurate and consistent translation of named entities which greatly impacts the factuality accuracy of news translation.

\bibliography{anthology,custom}
\bibliographystyle{acl_natbib}

\newpage
\appendix

\section{Multi-encoder architecture} \label{sec:appendix-multi-enc}
As been systematically explored by ~\citet{libovicky-etal-2018-input}, there are four kinds of multi-encoder cross-attention that can be applied on the decoder side: (1) Serial: cross-attention to each encoder is performed layer by layer. (2) Parallel: cross-attention to each encoder is performed in parallel and then the outputs are added together before feeding to the feed-forward layer. (3) Flat: outputs of all encoders are concatenated along the length dimension as the input to a single cross-attention. (4) Hierarchical: a second attention block is added to attend to the representations output by the parallel cross-attention.
While models in Table~\ref{tab:not-working-methods} all use the parallel cross-attention described in \S~\ref{sec:input-combination}, Table~\ref{tab:multi_src_xattn} ablates different multi-source cross-attention mechanisms. Three out of four cross-attention achieve similar performance, whereas the `flat' attention is considerably worse. This echos the findings by ~\citet{libovicky-etal-2018-input}.

\begin{table}[!h]
    \centering
    \begin{tabular}{lclc}
    \toprule
       Config.  &  BLEU &  Config.  &  BLEU \\\midrule
       Serial & 34.1 & Flat & 24.9 \\
       Parallel & 34.0 & Hierarchical & 34.1\\ \bottomrule
    \end{tabular}
    \caption{Indic test set BLEU scores of multi-encoder architecture trained on \base+\transl\ using different multi-source cross-attention. All mechanisms perform similarly except \textit{flat} cross-attention.}
    \label{tab:multi_src_xattn}
\end{table}

\section{Experiments} \label{sec:appendix-experiments}

\subsection{Data statistics}
The number of training examples for each language in both Turkic and Indic dataset is shown in Table~\ref{tab:data_stats}. We evaluate the Turkic dataset on multi-way FloRes101 devtest set, each having 1012 examples. To evaluate the Indic models, we use the provided multi-way test set of WAT21 MultiIndic task, each having 2390 examples.


\subsection{Input concatenation analysis} \label{sec:appendix-input-concat}
\begin{table}[!h]
    \centering
    \scalebox{0.9}{
        \begin{tabular}{lc|lc}
    \toprule
    Config. & BLEU & Config. & BLEU \\ \midrule
        \base\ + \ipa & 23.3 & \ipa\ + \base & 23.0 \\
        \base' + \ipa & 3.3 & \ipa' + \base & 13.9\\
        \base\ + \ipa' & 20.2 & \ipa\ + \base' & 9.5\\
    \bottomrule
    \end{tabular}
    }
    \caption{Models trained on concatenated original and phonetic input while evaluated on partially corrupted input. We use \ipa' to denote the phonetic part of the input is in corruption. Results are reported on the FloRes101 Indic languages instead of MultiIndic test set.}
    \label{tab:input-concat-ablate}
\end{table}

In \S~\ref{sec:results}, results show that models trained on the concatenated input does not bring any discernible improvement, but rather the performance is almost the same. To understand if the model has indeed utilized the concatenated alternative signals, we take the trained model and evaluate BLEU scores on the corrupted input. Specifically, one part of the concatenated input is corrupted while the other is left intact. The corruption is done by shuffling the tokens within the selected part of the input. Overall, we find that the model indeed pays attention to both parts of the input, as corrupting any part of them leads to large regression in BLEU scores (Table~\ref{tab:input-concat-ablate}). Moreover, no matter which type of signal is put in the front of the sentence, the model always pays more attention to the original input rather than the phonetic input, since corrupting the original input causes larger performance degradation than corrupting the phonetic input.

\begin{figure*}
\centering
    \includegraphics[width=\textwidth]{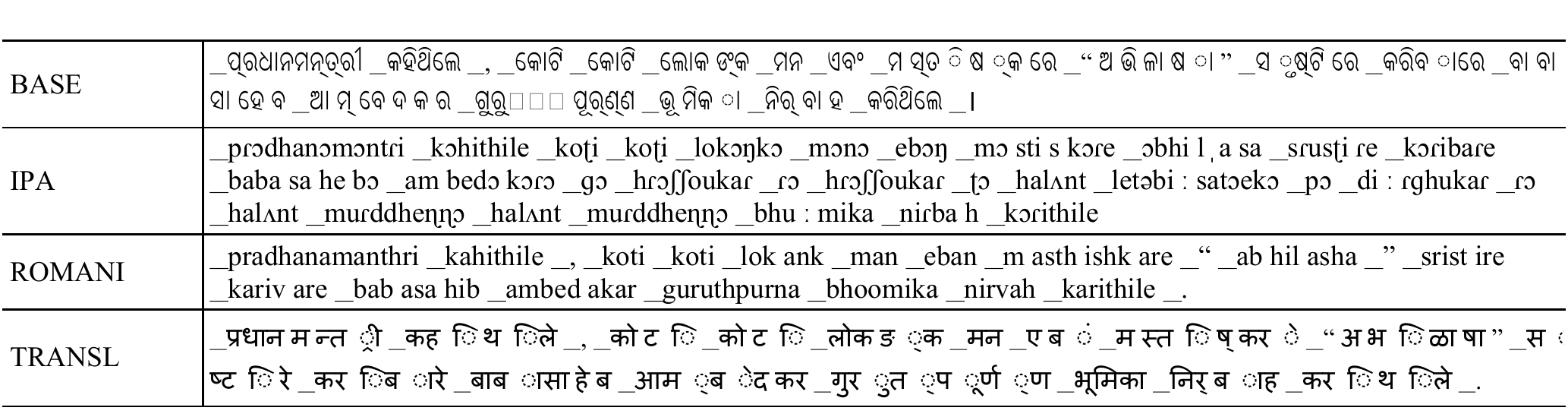}
    \caption{Example alternative signals. \textbf{BASE} is the original input in Oriya script, \textbf{IPA} is the phonetic input, \textbf{ROMANI} the romanized input, and \textbf{TRANSL} (\transl\ in the main text) the input transliterated into Devanagari script.}
    \label{fig:example_signals}
\end{figure*}

\section{Example alternative input signal}
\label{sec:appendix-signal-example}
\begin{figure*}
    \centering
    \includegraphics[width=\textwidth]{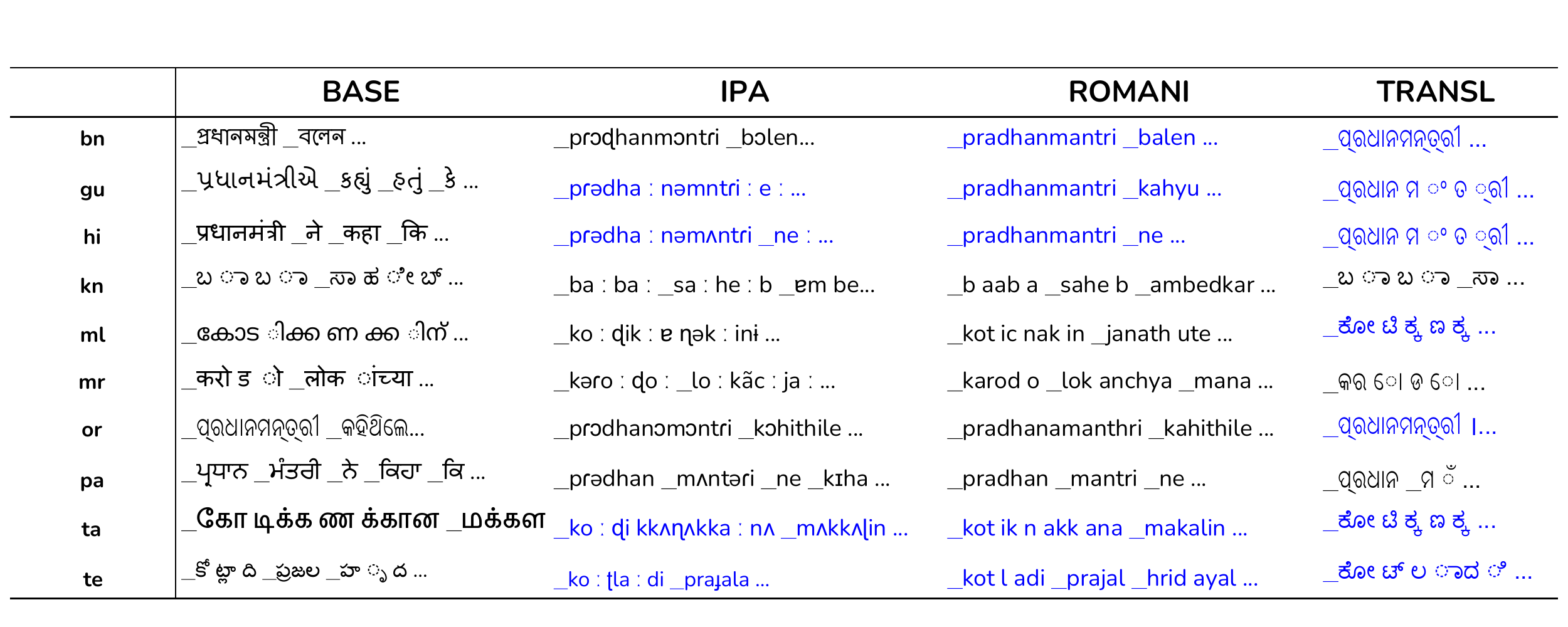}
    \caption{Example alternative signals of the same sentence in ten Indic languages. The token overlap across multiple languages are highlighted in blue. Compared to the original input, transliteration significantly increases token overlap.}
    \label{fig:alter-signals}
\end{figure*}

We present example alternative signals in Figure~\ref{fig:example_signals} and Figure~\ref{fig:alter-signals}. When the input are transformed to scripts other than their native script, there are more shared tokens in the source languages (as highlighted in Figure~\ref{fig:alter-signals}).

\begin{table*}[ht]
\centering
\begin{tabular}{lc|lclc}
\toprule
\multicolumn{2}{c|}{\textbf{Turkic languages}}            & \multicolumn{4}{c}{\textbf{Indic Languages}}                                                       \\ \midrule
\textbf{Language} & \textbf{\#bi-text} & \textbf{Language} & \textbf{\#bi-text} & \hspace{0.5cm} \textbf{Language} & \textbf{\#bi-text} \\ \midrule
Kazakh            & 919,877            & Bengali           & 1,756,197          & \hspace{0.5cm} Marathi           & 781,872            \\
Kyrgyz            & 243,179            & Gujarati          & 518,015            & \hspace{0.5cm} Oriya             & 252,160            \\
Turkish           & 4,000,000          & Hindi             & 3,534,387          & \hspace{0.5cm} Punjabi           & 518,508            \\
Uzbek             & 156,615            & Kannada           & 396,865            & \hspace{0.5cm} Tamil             & 1,499,441          \\
Azerbaijani       & 1,847,723          & Malayalam         & 1,204,503          & \hspace{0.5cm} Telugu            & 686,626            \\ \bottomrule
\end{tabular}
\caption{Training data statistics for Turkic and Indic dataset.}
\label{tab:data_stats}
\end{table*}

\section{Analysis}

\subsection{Token overlap details}

In \S~\ref{sec:results} we show the token overlap of various signals aggregated over all source language pairs, in this section we show the token overlap of each source language pair in Table~\ref{tab:tok-overlap-base} for the original input and in Table~\ref{tab:tok-overlap-transl} for the in-family transliterated input\footnote{Target script for Indo-Aryan languages is Oriya and Dravidian languages Kannada.}. Before performing transliteration, all source languages share only a small amount of token overlap except between Marathi and Hindi. The shared tokens between native scripts are mostly punctuation marks, digits and English tokens.  After transliteration, the token overlap becomes more obvious and a clear division between language families can be found. 

\begin{table}[ht]
    \centering
    \scalebox{0.65}{
    \begin{tabular}{l|rrrrrrrrrr}
\hline
   & \multicolumn{1}{l}{bn}       & \multicolumn{1}{l}{hi}       & \multicolumn{1}{l}{pa}       & \multicolumn{1}{l}{or}       & \multicolumn{1}{l}{gu}       & \multicolumn{1}{l}{mr}       & \multicolumn{1}{l}{kn}       & \multicolumn{1}{l}{ml}       & \multicolumn{1}{l}{ta}       & \multicolumn{1}{l}{te}       \\ \hline
bn & \multicolumn{1}{l}{}         & \cellcolor[HTML]{FFF6DB}0.05 & \cellcolor[HTML]{FFF8E4}0.04 & \cellcolor[HTML]{FFF8E4}0.04 & \cellcolor[HTML]{FFFFFF}0.01 & \cellcolor[HTML]{FFFFFF}0.01 & \cellcolor[HTML]{FFFFFF}0.01 & \cellcolor[HTML]{FFFFFF}0.01 & \cellcolor[HTML]{FFFFFF}0.01 & \cellcolor[HTML]{FFFFFF}0.01 \\
hi & \cellcolor[HTML]{FFF6DB}0.05 & \multicolumn{1}{l}{}         & \cellcolor[HTML]{FFF3D2}0.06 & \cellcolor[HTML]{FFF6DB}0.05 & \cellcolor[HTML]{FFFDF6}0.02 & \cellcolor[HTML]{FFD666}0.18 & \cellcolor[HTML]{FFFFFF}0.01 & \cellcolor[HTML]{FFFFFF}0.01 & \cellcolor[HTML]{FFFDF6}0.02 & \cellcolor[HTML]{FFFFFF}0.01 \\
pa & \cellcolor[HTML]{FFF8E4}0.04 & \cellcolor[HTML]{FFF3D2}0.06 & \multicolumn{1}{l}{}         & \cellcolor[HTML]{FFF8E4}0.04 & \cellcolor[HTML]{FFFDF6}0.02 & \cellcolor[HTML]{FFFDF6}0.02 & \cellcolor[HTML]{FFFFFF}0.01 & \cellcolor[HTML]{FFFFFF}0.01 & \cellcolor[HTML]{FFFDF6}0.02 & \cellcolor[HTML]{FFFFFF}0.01 \\
or & \cellcolor[HTML]{FFF8E4}0.04 & \cellcolor[HTML]{FFF6DB}0.05 & \cellcolor[HTML]{FFF8E4}0.04 & \multicolumn{1}{l}{}         & \cellcolor[HTML]{FFFFFF}0.01 & \cellcolor[HTML]{FFFFFF}0.01 & \cellcolor[HTML]{FFFFFF}0.01 & \cellcolor[HTML]{FFFFFF}0.01 & \cellcolor[HTML]{FFFFFF}0.01 & \cellcolor[HTML]{FFFFFF}0.01 \\
gu & \cellcolor[HTML]{FFFFFF}0.01 & \cellcolor[HTML]{FFFDF6}0.02 & \cellcolor[HTML]{FFFDF6}0.02 & \cellcolor[HTML]{FFFFFF}0.01 & \multicolumn{1}{l}{}         & \cellcolor[HTML]{FFF6DB}0.05 & \cellcolor[HTML]{FFF8E4}0.04 & \cellcolor[HTML]{FFF8E4}0.04 & \cellcolor[HTML]{FFF6DB}0.05 & \cellcolor[HTML]{FFF8E4}0.04 \\
mr & \cellcolor[HTML]{FFFFFF}0.01 & \cellcolor[HTML]{FFD666}0.18 & \cellcolor[HTML]{FFFDF6}0.02 & \cellcolor[HTML]{FFFFFF}0.01 & \cellcolor[HTML]{FFF6DB}0.05 & \multicolumn{1}{l}{}         & \cellcolor[HTML]{FFF8E4}0.04 & \cellcolor[HTML]{FFF6DB}0.05 & \cellcolor[HTML]{FFF6DB}0.05 & \cellcolor[HTML]{FFF8E4}0.04 \\
kn & \cellcolor[HTML]{FFFFFF}0.01 & \cellcolor[HTML]{FFFFFF}0.01 & \cellcolor[HTML]{FFFFFF}0.01 & \cellcolor[HTML]{FFFFFF}0.01 & \cellcolor[HTML]{FFF8E4}0.04 & \cellcolor[HTML]{FFF8E4}0.04 & \multicolumn{1}{l}{}         & \cellcolor[HTML]{FFF8E4}0.04 & \cellcolor[HTML]{FFF6DB}0.05 & \cellcolor[HTML]{FFF8E4}0.04 \\
ml & \cellcolor[HTML]{FFFFFF}0.01 & \cellcolor[HTML]{FFFFFF}0.01 & \cellcolor[HTML]{FFFFFF}0.01 & \cellcolor[HTML]{FFFFFF}0.01 & \cellcolor[HTML]{FFF8E4}0.04 & \cellcolor[HTML]{FFF6DB}0.05 & \cellcolor[HTML]{FFF8E4}0.04 & \multicolumn{1}{l}{}         & \cellcolor[HTML]{FFF6DB}0.05 & \cellcolor[HTML]{FFF8E4}0.04 \\
ta & \cellcolor[HTML]{FFFFFF}0.01 & \cellcolor[HTML]{FFFDF6}0.02 & \cellcolor[HTML]{FFFDF6}0.02 & \cellcolor[HTML]{FFFFFF}0.01 & \cellcolor[HTML]{FFF6DB}0.05 & \cellcolor[HTML]{FFF6DB}0.05 & \cellcolor[HTML]{FFF6DB}0.05 & \cellcolor[HTML]{FFF6DB}0.05 & \multicolumn{1}{l}{}         & \cellcolor[HTML]{FFF6DB}0.05 \\
te & \cellcolor[HTML]{FFFFFF}0.01 & \cellcolor[HTML]{FFFFFF}0.01 & \cellcolor[HTML]{FFFFFF}0.01 & \cellcolor[HTML]{FFFFFF}0.01 & \cellcolor[HTML]{FFF8E4}0.04 & \cellcolor[HTML]{FFF8E4}0.04 & \cellcolor[HTML]{FFF8E4}0.04 & \cellcolor[HTML]{FFF8E4}0.04 & \cellcolor[HTML]{FFF6DB}0.05 & \multicolumn{1}{l}{}         \\ \hline
\end{tabular}
    }
    \caption{Token overlap \base}
    \label{tab:tok-overlap-base}
\end{table}

\begin{table}[ht]
    \centering
    \scalebox{0.65}{
    \begin{tabular}{l|rrrrrrrrrr}
\hline
   & \multicolumn{1}{l}{bn}       & \multicolumn{1}{l}{hi}       & \multicolumn{1}{l}{pa}       & \multicolumn{1}{l}{or}       & \multicolumn{1}{l}{gu}       & \multicolumn{1}{l}{mr}       & \multicolumn{1}{l}{kn}       & \multicolumn{1}{l}{ml}       & \multicolumn{1}{l}{ta}       & \multicolumn{1}{l}{te}       \\ \hline
bn & \multicolumn{1}{l}{}         & \cellcolor[HTML]{FFE499}0.33 & \cellcolor[HTML]{FFEAB0}0.26 & \cellcolor[HTML]{FFE8A6}0.29 & \cellcolor[HTML]{FFE59D}0.32 & \cellcolor[HTML]{FFE8A6}0.29 & \cellcolor[HTML]{FFFEF9}0.03 & \cellcolor[HTML]{FFFEF9}0.03 & \cellcolor[HTML]{FFFEF9}0.03 & \cellcolor[HTML]{FFFEF9}0.03 \\
hi & \cellcolor[HTML]{FFE499}0.33 & \multicolumn{1}{l}{}         & \cellcolor[HTML]{FFD666}0.49 & \cellcolor[HTML]{FFEAB0}0.26 & \cellcolor[HTML]{FFD666}0.49 & \cellcolor[HTML]{FFDE83}0.4  & \cellcolor[HTML]{FFFEF9}0.03 & \cellcolor[HTML]{FFFEF9}0.03 & \cellcolor[HTML]{FFFEF9}0.03 & \cellcolor[HTML]{FFFEF9}0.03 \\
pa & \cellcolor[HTML]{FFEAB0}0.26 & \cellcolor[HTML]{FFD666}0.49 & \multicolumn{1}{l}{}         & \cellcolor[HTML]{FFF0C6}0.19 & \cellcolor[HTML]{FFE18D}0.37 & \cellcolor[HTML]{FFE396}0.34 & \cellcolor[HTML]{FFFEF9}0.03 & \cellcolor[HTML]{FFFEF9}0.03 & \cellcolor[HTML]{FFFEF9}0.03 & \cellcolor[HTML]{FFFEF9}0.03 \\
or & \cellcolor[HTML]{FFE8A6}0.29 & \cellcolor[HTML]{FFEAB0}0.26 & \cellcolor[HTML]{FFF0C6}0.19 & \multicolumn{1}{l}{}         & \cellcolor[HTML]{FFEDB9}0.23 & \cellcolor[HTML]{FFEEBD}0.22 & \cellcolor[HTML]{FFFFFF}0.01 & \cellcolor[HTML]{FFFFFF}0.01 & \cellcolor[HTML]{FFFFFF}0.01 & \cellcolor[HTML]{FFFFFF}0.01 \\
gu & \cellcolor[HTML]{FFE59D}0.32 & \cellcolor[HTML]{FFD666}0.49 & \cellcolor[HTML]{FFE18D}0.37 & \cellcolor[HTML]{FFEDB9}0.23 & \multicolumn{1}{l}{}         & \cellcolor[HTML]{FFDD80}0.41 & \cellcolor[HTML]{FFFEF9}0.03 & \cellcolor[HTML]{FFFEF9}0.03 & \cellcolor[HTML]{FFFEF9}0.03 & \cellcolor[HTML]{FFFEF9}0.03 \\
mr & \cellcolor[HTML]{FFE8A6}0.29 & \cellcolor[HTML]{FFDE83}0.4  & \cellcolor[HTML]{FFE396}0.34 & \cellcolor[HTML]{FFEEBD}0.22 & \cellcolor[HTML]{FFDD80}0.41 & \multicolumn{1}{l}{}         & \cellcolor[HTML]{FFFEF9}0.03 & \cellcolor[HTML]{FFFEF9}0.03 & \cellcolor[HTML]{FFFEF9}0.03 & \cellcolor[HTML]{FFFEF9}0.03 \\
kn & \cellcolor[HTML]{FFFEF9}0.03 & \cellcolor[HTML]{FFFEF9}0.03 & \cellcolor[HTML]{FFFEF9}0.03 & \cellcolor[HTML]{FFFFFF}0.01 & \cellcolor[HTML]{FFFEF9}0.03 & \cellcolor[HTML]{FFFEF9}0.03 & \multicolumn{1}{l}{}         & \cellcolor[HTML]{FFEBB3}0.25 & \cellcolor[HTML]{FFF2CC}0.17 & \cellcolor[HTML]{FFE6A0}0.31 \\
ml & \cellcolor[HTML]{FFFEF9}0.03 & \cellcolor[HTML]{FFFEF9}0.03 & \cellcolor[HTML]{FFFEF9}0.03 & \cellcolor[HTML]{FFFFFF}0.01 & \cellcolor[HTML]{FFFEF9}0.03 & \cellcolor[HTML]{FFFEF9}0.03 & \cellcolor[HTML]{FFEBB3}0.25 & \multicolumn{1}{l}{}         & \cellcolor[HTML]{FFE293}0.35 & \cellcolor[HTML]{FFE499}0.33 \\
ta & \cellcolor[HTML]{FFFEF9}0.03 & \cellcolor[HTML]{FFFEF9}0.03 & \cellcolor[HTML]{FFFEF9}0.03 & \cellcolor[HTML]{FFFFFF}0.01 & \cellcolor[HTML]{FFFEF9}0.03 & \cellcolor[HTML]{FFFEF9}0.03 & \cellcolor[HTML]{FFF2CC}0.17 & \cellcolor[HTML]{FFE293}0.35 & \multicolumn{1}{l}{}         & \cellcolor[HTML]{FFEAB0}0.26 \\
te & \cellcolor[HTML]{FFFEF9}0.03 & \cellcolor[HTML]{FFFEF9}0.03 & \cellcolor[HTML]{FFFEF9}0.03 & \cellcolor[HTML]{FFFFFF}0.01 & \cellcolor[HTML]{FFFEF9}0.03 & \cellcolor[HTML]{FFFEF9}0.03 & \cellcolor[HTML]{FFE6A0}0.31 & \cellcolor[HTML]{FFE499}0.33 & \cellcolor[HTML]{FFEAB0}0.26 & \multicolumn{1}{l}{}         \\ \hline
\end{tabular}
    }
    \caption{Token overlap \transl}
    \label{tab:tok-overlap-transl}
\end{table}

\subsection{Similarity in latent space}

Besides examining the consistency of system output as in \S~\ref{sec:consist-output}, we also measure the distance of source representations in the latent space. Concretely, we compute the average of normalized Euclidean distance over all source language pairs: 
\[\frac{1}{{N \choose 2}}\sum_{m, n}dist(l_m, l_n)\]
, where $N$ is the total number of source languages, $dist(l_m, l_n)$ compute the distance between a sentence in language $m$ and language $n$:

\begin{align*}
     dist( & l_m, l_n) = \\
    & \frac{1}{2} \big( \frac{1}{|l_m|}\sum_i \min_j\big(dist(w_{mi}, w_{nj}) \big) + \\
    & \frac{1}{|l_n|} \sum_j \min_i\big(dist(w_{mi}, w_{nj}) \big) \big)
\end{align*}

, where $|l_m|$ and $|l_n|$ are the number of tokens within sentence in language $m$ and language $n$ respectively. $w_{mi}$ represents the $i^{th}$ encoder output of sentence $l_m$. $dist$ represents the Euclidean distance between the two vectors. We additionally normalize the distance with $\sqrt{d}$ where $d$ is the dimension of the dense vector. Adding the scaling factor is to make the scaled self-ensemble model comparable with the rest variants. 

As shown in Table~\ref{tab:repr-dist}, none of the alternative signals alone can lead to more similar source representations. While training on the original input and one alternative input, only the combination of \base\ and \transl\ lowers the distance of original input representations from 0.60 to 0.58. The distances become even more smaller while training the scaled self-ensemble model. The distances among \base\ representations decrease to 0.54 and the rest three input signals all yield more similar representations than the original input. Overall, we didn't find significant differences in latent space, which we would like to keep investigating in the future.

\begin{table}[]
    \centering
    \scalebox{0.75}{
        \begin{tabular}{@{}lcccc@{}}
\toprule
Config.                      & \base & \ipa  & 
\romani & \transl \\ \midrule
Trained separately           & 0.60 & 0.62 & 0.60   & 0.61   \\
SE(\base+\ipa)                 & 0.60 & 0.62 & -      & -      \\
SE(\base+\romani)              & 0.61 & -    & 0.60   & -      \\
SE(\base+\transl)              & 0.58 & -    & -      & 0.60   \\
SE(ALL)   & 0.60 & 0.62 & 0.60   & 0.61   \\
S-SE(ALL) & 0.54 & 0.53 & 0.52   & 0.52   \\ \bottomrule
\end{tabular}
    }
    \caption{Normalized Euclidean distances of single-input model (Trained separately), self-ensemble model (SE) and scaled self-ensemble model (S-SE).}
    \label{tab:repr-dist}
\end{table}

\end{document}